# Assessing Partisan Traits of News Text Attributions During the 2016 US Presidential Election

**Logan Martel**
McGill University
Computer Science

**Edward Newell**
McGill University
Computer Science

**Drew Margolin**
Cornell University
Dept. of Communication

**Derek Ruths**
McGill University
Computer Science

{logan.martel, edward.newell}@mail.mcgill.ca,
dm658@cornell.edu, derek.ruths@mcgill.ca

## Abstract

On the topic of journalistic integrity, the current state of accurate, impartial news reporting has garnered much debate in context to the 2016 US Presidential Election. In pursuit of computational evaluation of news text, the statements (*attributions*) ascribed by media outlets to *sources* provide a common category of evidence on which to operate. In this paper, we develop an approach to compare partisan traits of news text attributions and apply it to characterize differences in statements ascribed to candidate, Hilary Clinton, and incumbent President, Donald Trump. In doing so, we present a model trained on over 600 in-house annotated attributions to identify each candidate with accuracy > 88%. Finally, we discuss insights from its performance for future research.

## 1 Introduction

Laying the groundwork for computational comparison of statements attributed to Mr. Trump and Mrs. Clinton, prior research (Newell, Schang, Margolin, & Ruths, 2017) has operationalized attributions according to the PARC3 annotation format (Silvia Pareti, 2015, 2016). Here, an attribution is defined as the 3-tuple (**S, C, T**) = (**source, cue, content**)[1].

Intuitively, this structured description of an attribution captures the notion that *attributions* comprise not only *what* (**content**) was attributed, but also *who* (**source**) said it, and *how* (**cue**) that source was linked to the content. More precisely, we can say that an attribution occurs in text when a statement conveys "the discourse, attitude, or inner state of an external source" (Piazza, 2009). In the scope of journalism, attributions are often leveraged to certify the assertions of a news article (Reich, 2010), and act as "objective" evidence to the stances and circumstances being reported (Esser & Umbricht, 2014).

Under this lens, it is easy to see how attributions might be used as a mechanism to implicitly represent the character of public figures – in cases both where the public figure is either the source or subject of an attribution statement. For example, in a Times article reporting on "All the Times Donald Trump Insulted Mexico"[2], we see the following attribution:

> "They are not our friend, believe me," **he said, before disparaging Mexican immigrants:** "They're bringing drugs. They're bringing crime. They're rapists. And some, I assume, are good people."

In this case, our attribution statement begins immediately after the first quote, with the source "**he**" referring to Donald Trump. Evidently, in attributing the quote to Donald Trump, the Times article leverages the attribution as evidence to support its assertion that Trump has indeed "Insulted Mexico".

---

[1] Which shall be decomposed in detail in later

[2] http://time.com/4473972/donald-trump-mexico-meeting-insult/

Similarly, we see the same utility of an attribution in effect when a public figure is the subject of an attribution, rather than its source. For instance, consider a CNN report headlined that Bernie Sander's supporters have begun "to coalesce around Clinton"[3]. Here, we see an attribution ascribed to the Communications Workers of America (previously reported to have officially backed Sanders[4]) that:

> **the union** touted Clinton as "*thoughtful and experienced*" and thanked her for standing "*with CWA members and pledg[ing] her commitment to making life better for working families.*"

With this attribution, the article directly demonstrates its allegation that prior Sanders supporters are now beginning to support Mrs. Clinton. Moreover, independent of the article context, the attribution alone clearly has sufficient detail to convey a positive stance towards Hilary Clinton. In this sense, we can say that the attribution *positively represents* Hilary Clinton. Conversely, irrespective of underlying intent, by serving as evidence to the claim that Trump has "Insulted Mexico", we can intuitively say that the Time's attribution *negatively represents* Donald Trump.

Rooted in this intuitive notion that attributions can characterize some *positive* or *negative* representation of a subject entity, we arrive at a basis for comparison of attributions. Of course, in pursuing this notion, it is essential to be wary of overemphasizing its implications on a case-by-case basis. That attributions positively or negatively represent a subject in any given article is far from sufficient evidence for an underlying partisan lean or personal bias of the text's author. In fact, on evaluation of the "objectivity" of a journalistic work, we would actually expect a wide range of "facts and opinions that conflict" (Ryan, 2001), comprising many instances of attributions that both positively and negatively represent a named entity.

Rather, in formulating a more rigorous standard for comparison, we should be surprised only to observe large differentials in the frequency of attributions positively or negatively representing a given subject. To this end, we require a formal model to amalgamate meaningful partisan traits of attributions. Thus, our work has demonstrated success in further decomposing attributions and performing computational analysis to identify differences across diverse attribution categories.

As part of our approach, we build a dataset[5] of partisan-trait annotated attributions, as an additional layer on top of news text previously-annotated by both Penn Treebank P.O.S. tags (Marcus, Santorini, & Marcinkiewicz, 1993) and the Penn Attribution Relations Corpus, version 3 (PARC3) (S. Pareti, 2012; Silvia Pareti, 2015, 2016). By design, our attributions dataset consists of statements ascribed to Hilary Clinton and Donald Trump respectively, in addition to other candidates and non-candidate actors, sampled in approximately equal measure across over 25 articles and 7 media outlets.

Driven by this annotated data, we conduct statistical analysis on *ad-hoc* hypothesized structure in our attributions. Moreover, we leverage insights from our analysis to identify systemic challenges when investigating political attributions and demonstrate results in addressing some of these challenges. Most tangibly, we implement a classifier to identify our target candidates with high accuracy and discuss its immediate applicability at scale in future research.

## 2 Related Work

### 2.1 Computational analysis of attributions

This research effort builds fundamentally on prior efforts to architect the analysis of news text attributions as an end-to-end system. Most notably, our system directly leverages and extends from open-source contributions to processing verifiability-scored PARC3 attributions (Newell et al., 2017).

Predating these recent system implementations, early efforts tackled attributions fundamentally as the problem of extracting attributed content (e.g. a direct quote) and identifying the source of that quote. Early algorithms were powered by rules-based inference engines (Mamede & Chaleira, 2015; Zhang, Black, & Sproat, 2003) – enacted to classify speakers in children stories. These rules-based systems excelled at quote extraction but performed poorly in classifying the correct speakers (especially when tested across several genres).

---

[3] https://www.cnn.com/2016/07/11/politics/sanders-supporters-endorse-clinton-for-president/index.html

[4] https://www.cwa-union.org/news/entry/cwa_endorses_sen_bernie_sanders_for_president

[5] [Insert Link]

Evidently, the challenge of news text attribution extraction is further obfuscated by the prevalence of informal reporting practices (Baym, 2005) broadening the scope far beyond simple parsing of consistently-quote-enclosed character strings. Prior to major recent advancements in modern machine learning techniques (Qiu, Wu, Ding, Xu, & Feng, 2016), low recall pervaded as a tradeoff for high precision in parsing and sourcing attributions (Pouliquen, Steinberger, & Best, 2007; Sarmento & Nunes, 2009). More recently, machine learning approaches for attribution extraction have demonstrated effective performance (O'Keefe, Pareti, Curran, Koprinska, & Honnibal, 2012) by a sequence-labelling approach leveraging Penn P.O.S. tags and the Penn Discourse TreeBank (Miltsakaki, Prasad, Joshi, & Webber, 2004; Prasad et al., 2008) to learn from text annotations.

In stride with these recent advancements, efforts have been made to formalize attribution annotations. Of principle note, the PARC3 annotation format (Silvia Pareti, 2015, 2016) specifies the fundamental input structure underlying both the computational approaches developed here, and in the open-source attribution systems which we extend (Newell et al., 2017). As introduced earlier, the PARC3 format specifies attributions as the 3-tuple (**S, C, T**) = (**source, cue, content**). For completeness, we define: **S** as the source to whom the content is attributed, **C** as the cue phrase (e.g. *said*, *argued*) – typically a verb indicating how the source is linked to the content, and **T** as the content being attributed – typically an assertion or quote about some target subject. Aggregating a dataset of these PARC3 attributions across several distinct articles and media outlets, our research proceeds then by first decomposing further dimensions of attributions for analysis, crowdsourcing annotations on these dimensions, and modelling the result.

### 2.2 Features of Attributions in News Text

Attributions have been described as the "bread and butter" of hard news journalism (Sundar, 1998). In this context, it has been said that "external voices are allowed to speak their mind much more loudly than journalists"(Jullian, 2011). Evidently, in the spirit of our earlier examples of attributions positively and negatively representing a target subject, this mechanism enacts attributions as tool to "frame" (Johnson-Cartee, 2004) the narratives being reported by proxy. The use of attributions in this fashion is often subtle (Wodak & Fairclough, 2013).

For our purposes, we are most interested in modelling the features that underlie this enactment of attributions as apparatuses for narrative framing. Thus far, a big focus in this domain has centered on the construction and application of validation datasets, such as CREDBANK (Mitra & Gilbert, 2015) to measure the credibility of event mentions attributed to sources. Recently, early work in this direction has achieved success in predicting the veridicality (ie. "factuality") of event mentions found in quoted source-introducing predicates on Twitter (Soni, Mitra, Gilbert, & Eisenstein, 2014) by regressing on cue-words lemmatized to hard-coded factuality indicators. Among other patterns in quote variation, frequently recurring linguistic phenomenon (Lauf, Valette, & Khouas, 2013) include synonymic variations in verb modifiers, patterns of coreference, and modality.

A common thread that measures of emotive judgements arise as predictors for source credibility across these research efforts appears to suggest that personal stances of the source towards some target are a strong implicit feature of attribution statements. Consequently, it should come as no surprise then that research efforts in natural language processing are seeing an increasing distinction between *stance* and *sentiment* in subjectivity analysis (Montoyo, MartíNez-Barco, & Balahur, 2012). This distinction shall also lie at the core of our efforts to investigate partisan traits of the stances represented in attributions sourced from our 7 media outlets.

### 2.3 Partisan Traits of Attributions

As a first step towards modelling implicit features of attributions as predictors for partisan stances of the author, we again consider the role of the journalist in framing a narrative via attributions (Johnson-Cartee, 2004). Historically, efforts to quantify media bias (or *slant*) have looked primarily at characteristics on an article-by-article basis (Covert & Wasburn, 2007; D'Alessio & Allen, 2000; Tankard, 2001). Here, we seek a more granular approach focused on how the *style*, *structure*,

*sentiment*, and *stance* of attributions vary with respect to the reporting media outlet, and whether the attribution is sourced to Donald Trump, Hilary Clinton, or otherwise. For consistency, henceforth, we will refer to this group of characteristics as *partisan traits* of attributions.

## 3 Operationalizing Partisan Traits

### 3.1 Task Definition

The task of operationalizing partisan attribution traits, in preparation for further analysis, shall be defined as the processes required to construct a dataset of attributions labelled with sufficient information to make comparisons with regards to *style*, *structure*, *sentiment*, and *stance* between distinct categories of attributions. To achieve this goal, we break down our pipeline into the following phases: (1) sourcing data, (2) pre-processing the data, (3) annotating the data, (4) formatting the data for analysis.

For the sake of approximating value judgments of the general public, we execute the data annotation via a small cohort of human annotators. Annotators were each responsible separately for annotating and reviewing small subsets of random articles, given access to both the raw text of the article, and a pre-processed set of attribution 3-tuples known to be contained in that article. After formatting the human-annotated data (as of phase (4)), it is then ready both for direct statistical analysis on the annotations, and for further computational modelling of relationships in our attributions.

### 3.2 Sourcing the Data

Any attempt to model partisan characteristics in journalistic text of course requires a representative sample of politically-oriented news articles. Given this study's focus on presidential candidates, Donald J. Trump and Hilary Rodham Clinton, our sample criteria thus required a sizeable collection of political reports by several media outlets restricted to the election campaign period (defined, here, as November 8, 2015 to November 8, 2016). For the purposes of investigating our hypothesized partisan slant, seven media outlets were chosen to reflect a range of perceived partisan biases (as rated by the AllSides Bias Rating[6]) and embody both large and smaller scale publishers (distinguished courtesy of the Alliance for Audited Media[7]). Ultimately, the seven publishers reflected in this study are: The New York Times, USA Today, The Washington Post, Huffington Post, Breitbart, Politico, and The Western Journalism Center.

In programmatically sourcing data for these publishers, the need arises for a web crawler to fetch articles for the desired date range and topical criteria. Fortunately, prior research at the McGill Network Dynamics Lab[8], implementing the aforementioned web crawler, has made the required dataset available for the purposes of this research effort. Specifically, this collection entails over three hundred raw text articles, published up to one year after November 8, 2015, and curated to fulfill requirements for political subject matter and a high count of in-text matches for the individuals Hilary Clinton, Donald Trump, Bernie Sanders, and Ted Cruz.

### 3.3 Preprocessing the Data

Accompanied with the raw text corpus, each article has additionally been tokenized to produce an auxiliary XML file via Stanford's CoreNLP pipeline (Manning et al., 2014), and is furthermore associated to a proprietary "BRATT" file enumerating attributions in the raw text. Another courtesy of prior work at the McGill Network Dynamics Lab, each BRATT file embodies a crowd-sourcing effort to identify all in-text attributions according to the consumable PARC3 (*source*, *cue*, *content*) attribution format (Silvia Pareti, 2015, 2016). For the sake of notational convenience, we shall then henceforth define a *consolidated_article* as the 3-tuple (*raw_text*, *core_nlp_tokens*, *bratt_attributions*) where each component refers to the respective data corpus described above.

Commencing our research from the above baseline, the following pre-processing tasks remained: (i) consolidate the raw text, CoreNLP, and BRATT data per-article to produce a collection of *consolidated_articles*, and (ii) of several-hundred *consolidated_articles*, identify a subset of 2-3 per publisher containing sufficient Trump and Clinton attributions for further analysis. Seeking to unite our data, the first task was eased by each article's unique mapping to an *article_key* of (*publisher_name*, *article_name*). From this collection of *consolidated_articles*, our next goal to identify the subset likely to contain Trump & Clinton attributions required a significant curation effort. Encompassing a key problem addressed by our research, the initial curation of this data leveraged both exploratory statistics (counting in-text matches for custom Trump/Clinton regular expressions), and manual scans of randomly-selected articles. For future replicability of this experiment, our initial articles collection (henceforth denoted *initial_validation_articles*) was leveraged in validating the performance of an in-house attribution-source classification algorithm readily deployable to seek "high-yield" articles for an input target candidate.

---

[6] https://www.allsides.com/media-bias/about-bias

[7] https://web.archive.org/web/20151016155148/http://auditedmedia.com/news/blog/top-25-us-newspapers-for-march-2013.aspx

[8] http://networkdynamics.org/

### 3.4 Annotating the Data

Given our set of *consolidated_articles* chosen for experimentation, the next step was to annotate each of our attributions to facilitate comparison of partisan characteristics. Towards this objective, a small cohort of human annotators was selected to assign labels to each of our attribution instances. Each annotator was initially assigned a subset of attributions from our *consolidated_articles* to label and, afterwards, assigned a subset of attributions, now labelled by other annotators, to review for inter-annotator agreement. Each annotator was then responsible for either reviewing or annotating attributions sourced from articles from at least 4 distinct publishers, and each annotator was involved at most once for attributions from a given media outlet.

To ensure consistency in annotation, the task was conducted according to the methodological corpus annotation guidelines prescribed by Eduard Hovid and Julia Lavid (Hovy & Lavid, 2010). Annotators were given an explicit Codebook detailing the available labels, with examples, for each annotation category, and vetted for annotating with accuracy >70% on a "training suite" article. Afterwards, each Annotator was given access to copies of their relevant *consolidated_articles* and, for each, instructed to (i) iterate over their assigned attributions in the article's *bratt_attributions* file, (ii) leverage the PARC (*source*, *cue*, *content*) breakdown for each attribution, and contextual information (as needed) from the raw text file to understand *who* and *what* is involved in each attribution, and finally (iii) produce labels for each attribution in a new *labelled_attributions* csv file. During review, any disagreement in annotation was discussed collectively to arrive at a consensus.

We shall now proceed to elaborate the details of our attribution labels, broken down by *source*, *cue*, and *content* features, in addition to *logistical* labels. In this endeavor, our goal is to design features that implicitly convey the *style*, *structure*, *sentiment*, and *stance* of attributions.

**Logistical Attribution Labels.** For future reference, each attribution is uniquely identified by a 3-tuple (3 label) *attribution_key* of the form (*publisher_name*, *article_name*, *attr_id*) where the first two components are as in an *article_key*, and *attr_id* is an index into the respective attribution as enumerated by each article's crowd-sourced BRATT file.

**Attribution Source Labels.** Of utmost importance in contrasting our political attributions, it is essential to know with confidence whether a given attribution was sourced to Hilary Clinton, Donald Trump, or otherwise. Given that an attribution's source text field may identify the source indirectly (e.g. "he/she said"), we take advantage of our human annotators to confidently de-reference and label our attributions via contextual information with one of several discrete *source_label* categories. It is easy to see how these "gold standard" labels are integral to validating both CoreNLP's vanilla co-reference resolution (Lee et al., 2013) and our in-house political candidate classifier.

Intuitively, in describing the source text of an attribution, two immediate stylistic observations are the *formality* and *tone* with which the source is cited. As one ubiquitous signal of *formality*, albeit more prominent in some languages than others, it has been said that "every language contains an *honorific* lexicon" (Agha, 1998). To this end, on identification of honorific tokens in our source field (e.g. "Mrs.", "The Right Honorable"), we tag the raw honorific string in our *honorific_text* label. Regarding *tone*, on consideration of cases where the source is cited with either reverence or disregard (e.g. "political mastermind Conway" vs. "crooked Hilary"), we denote *source_valence* to annotate the *positive*, *negative*, or *neutral* sentiment of an attribution's source field.

**Attribution Cue Labels.** On the basis that we might reasonably expect an arbitrarily large number of distinct cue labels (e.g. ranging from "said" to "confirmed" to "angrily protested"), it would seem ideal to restrict the scope of cue generalizations for our initial investigation. Rather, we explicitly tag only one cue-specific feature, *cue_valence*, which denotes (analogous to *source_valence*), the *positive*, *negative*, or *neutral* sentiment of an attribution's cue field.

**Attribution Content Labels.** Unsurprisingly, the scope of our attribution content labels is the broadest thus far by a considerable margin. Generally, while targeted principally at understanding the assertions conveyed by our attribution's content text, these labels are distinguished in that they often do require contextual information as conveyed by the attribution's *source* and *cue* fields (e.g. asserting that **Trump** lambasted, "…" affects interpretation of the "…" text). Breaking down the dimensions of our attribution content, we consider primarily its *stance*, *medium*, *directness*, and *attribution type*. Of the structural aspects of our attribution content, *medium* refers to the attribution's channel of origin (e.g. tweet, formal speech, etc.), and *directness*, despite intuitively occupying a continuum, is binarized for our purposes to indicate whether an attribution *is* or *is not* a direct quote. Further elaborating structural notions of our attributions, *attribution type* attempts to categorize attributions into one of the following ten categories: {*headline, political_platform, personal_stance, speech_snippet, trump_callout, clinton_callout, sanders_callout, cruz_callout, group_callout, other_callout*}. To summarize, these categories attempt to approximate mutual exclusivity by segregating special cases (e.g. *headline*, *speech_snippet*), ascribing neutral attributions as *callouts* (e.g. **Clinton** debated with *Trump* is a *trump_callout*), and differentiating *personal stances* (e.g. **Trump** promised to *"Make America Great Again"*) and *political platforms* (e.g. **Trump** promised to *lower taxes*).

Strongly related to the nature of these assertions, we also attempt to classify the *stance* of the attribution's **source** towards the **target** (ie. content text subject) of the assertion being attributed. The framework for this investigation emerges in our earlier, *related works*, discussion on the problem of *stance analysis*. Attracting considerable recent attention, the problem of *stance analysis* has been formalized (Mohammad, 2017) as involving both (a) identifying a (**source**, **target**) pair associated to the stance, and (b) classifying the *stance* of the **source** towards the **target** as one of: {*favours_target, against_target, neutral*}. Here, it is easy to see how *stance* fits as a dimension of attributions since it is always assigned to a **source** and the **target** can be any subject of discussion (e.g. a proposition, a named entity, or otherwise). Moreover, for our purposes, we are fortunate in that we can greatly simplify *stance analysis* by restricting our targets to only Donald Trump, Hilary Clinton, or Other. Thus, for our final content feature, we assign *stance_type* as the best-fitting label of the available annotations: {*favours_trump, favours_clinton, favours_other, against_trump, against_clinton, against_other, favours_both, against_both, neutral_both*}.

### 3.5 Formatting the Labelled Data

Now readily equipped with a collection of *labelled_articles*, some final preparation is required for programmatic consumption. First and foremost, we collect all *labelled_attribution* instances across each of our *labelled_articles* to a single csv dataset, henceforth denoted as the *labelled_attributions_dataset* or $D_A$. At the core of our analysis, programmatic consumption operates per-attribution by expecting an *attribution_key* of the form (*publisher_name*, *article_name*, *attr_id*) where the first two components are as in an *article_key*, and *attr_id* is an index into the respective attribution as enumerated by each article's crowd-sourced BRATT file. From this setup, we can then perform experiments by randomly sampling attribution instances from $D_A$ and indexing by *attribution_key* into the respective *consolidated_article* as needed. Integral to this effort, several open-source attributions research tools by Edward Newell[9] underlie the per-article data model. Indexing per article, any given attribution can then easily be associated to CoreNLP metadata (Manning et al., 2014) including tokenized POS tags, co-referent mentions, and dependency parses, as modelled by the corenlpy project[10].

## 4 Results

### 4.1 Preliminary Statistical Analysis

Immediately, aiming to quantitatively and qualitatively characterize differences in our Trump and Clinton attributions, we start by conducting hypothesis tests to contrast populations of interest and challenge our intuition on the contribution of the labelled attribution features.

Framing this analysis, there are two broad classes of comparisons: $source_1$ vs. $source_2$ (e.g. Trump vs. Clinton), and $publisher_1$ vs. $publisher_2$ (e.g. Breitbart vs. Huffpost). In conducting these tests, we randomly sample 100 attributions each (ie. 100 Trump vs. 100 Clinton) for source comparisons, and 75% of the minimum population size for publisher comparisons (ie. size(Breitbart, Huffpost) = (50, 40) => we sample floor(0.75 * 40) = 30 attributions each for Breitbart and Huffpost). Any composite sample (e.g. Trump-Huffpost) follows the same rules as in basic publisher comparisons.

For each population contrast, we conduct hypothesis tests for the significance of each of our attribution features and investigate basic models we had hypothesized *a priori* for interactions in these features. Typically, in testing basic feature significance, we construct a 2 x $L_i$ contingency table where the 2 rows correspond to our 2 contrast populations, and the $L_i$ columns are each of the discrete labels tagged to our attributions for the i-th feature of interest. Here, it is worth noting that we exclude labels for which the cell counts are 0, and apply either a $\chi^2$ Test of Independence (with Yate's Correction for Continuity) or Fisher's Exact Test as required for statistical validity (Agresti, 2013). Lastly, we model our contingency tables via Poisson Regressions characterizing functional dependencies to test for our hypothesized feature interactions.

The following table summarizes the experiments demonstrating sufficient evidence for feature contribution in discriminating our contrast populations at the $\alpha = 0.05$ significance level:

| Target_Population_Label | Contrast_Population_Label | Test_Type | Test_Factor | P_Value |
|---|---|---|---|---|
| trump | clinton | fisher_exact_test | honorific_text | 0.013234199 |
| trump | clinton | fisher_exact_test | stance_type | 4.80E-06 |
| trump | clinton | log_linear-chisq | stance-cue-val_interaction | 0.000210817 |
| trump | non_trump | fisher_exact_test | attr_type | 0.0036067 |
| trump | non_trump | fisher_exact_test | stance_type | 3.33E-05 |
| trump | non_trump | log_linear-chisq | stance-cue-val_interaction | 0.001837734 |
| trump | non_trump_or_clinton | chi_sq_test | cue_valence | 0.037025299 |
| trump | non_trump_or_clinton | fisher_exact_test | stance_type | 2.00E-07 |
| trump | non_trump_or_clinton | log_linear-chisq | stance-cue-val_interaction | 2.08E-05 |
| trump | non_trump_or_clinton | log_linear-chisq | stance-attr-type_interaction | 0.032234622 |
| clinton | non_clinton | fisher_exact_test | honorific_text | 0.016394898 |
| clinton | non_clinton | fisher_exact_test | attr_type | 0.0040192 |
| clinton | non_clinton | fisher_exact_test | stance_type | 5.90E-06 |
| clinton | non_clinton | log_linear-chisq | stance-cue-val_interaction | 0.000405472 |
| clinton | non_trump_or_clinton | fisher_exact_test | attr_type | 0.018609298 |
| clinton | non_trump_or_clinton | fisher_exact_test | stance_type | 0.000127 |
| clinton | non_trump_or_clinton | log_linear-chisq | stance-cue-val_interaction | 0.010286997 |
| clinton-breitbart | clinton-huffpost | fisher_exact_test | stance_type | 0.046625595 |
| trump-breitbart | clinton-breitbart | fisher_exact_test | stance_type | 0.018719498 |
| trump-huffpost | clinton-huffpost | fisher_exact_test | stance_type | 0.008525099 |

**Table 1:** Significant Hypothesis Test Results

Above, we see that only the following individual factors ever demonstrated evidence for a main effect: *honorific_text*, *stance_type*, *attr_type*, and *cue_valence*. Regarding *honorific_text*, further spot-checking unfortunately reveals that this manifested effectively only to discriminate on variations of the form "Mr." vs. "Mrs.".

On the other hand, we do see strong evidence for significance among the remaining three features. Of most consistent value, we are encouraged by the prevalence of *stance_type* across our population contrasts and, moreover, the recurrent evidence of strong interaction between *stance_type* and *cue_valence*. Tested independently, *cue_valence* demonstrated evidence for significance in discriminating between Trump and non-(Trump or Clinton) attributions. Similarly, the feature *attr_type* appears in several experiments to demonstrate its effectiveness at discriminating between attributions sourced to the primary political candidates (Trump or Clinton) and attributions sourced to anyone else. In particular, *stance_type* and *attr_type* evidenced strong interaction in a single experiment contrasting Trump and non-(Trump or Clinton) attributions.

---

[9] https://github.com/enewe101

[10] http://corenlpy.readthedocs.io/en/latest/

Aiming to better understand these features, all comparisons were additionally visualized by mosaic plots. Here, the frequency of our realized feature combinations is illustrated by the surface area of corresponding tiles, where the axes-dimensions are proportional to the marginal row/column frequencies. These visualizations are thus useful to confirm intuition on the qualitative contributions of our categories to differentiate attributions. As an example, consider the following plot involving Trump and Clinton attributions sourced from Huffington Post:

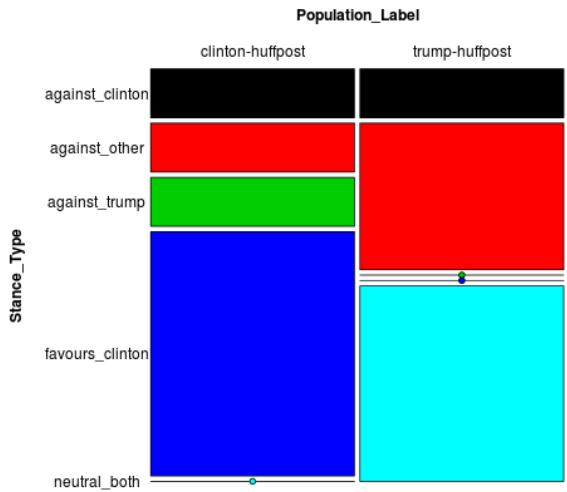

**Figure 1:** Huffpost Stance Types – Standard Mosaic Plot

At a glance, the above seems to forebode an interesting tendency for Huffpost to cite Clinton support herself and Trump making assertions against other targets. Digging deeper, we shade the tiles according to the magnitude of their Pearson residuals:

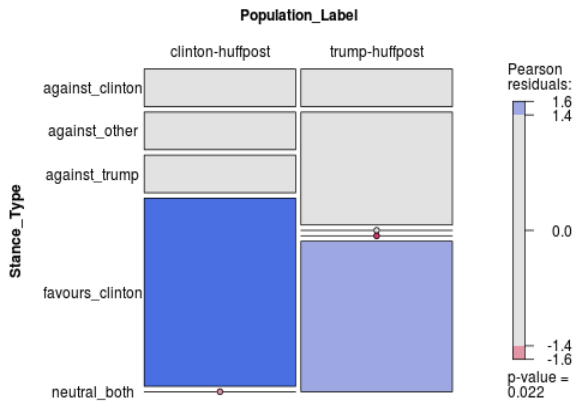

**Figure 2:** Huffpost Stance Types – Residuals Mosaic

Here, we interpret blue shading to color tiles that demonstrated more observations than expected under the null "feature independence" model, and red conversely shading tiles with fewer than expected counts. Thus, the above indeed illustrates evidence that Huffpost is significantly-often citing Clinton as favouring herself, however, we lack sufficient evidence to make the same claim about Trump attributions targeted against others.

Following the above approach to analyze our other contrast populations unearthed many other preliminary findings. Sanity-checking our most basic of expectations, we found that Clinton is never shown favouring Trump, Trump is never shown favouring Clinton, Clinton is often cited favouring herself, and Trump is similarly often shown supporting himself. More interestingly, across all 7 of our media outlets, we found ample evidence that Trump is often cited making claims *against non-Clinton targets* (moreso even than he makes claims directly against Clinton), but also that non-(Trump or Clinton) targets are most frequently cited in attributions targeted either *against Trump* or *favouring Clinton*. We illustrate this trend below:

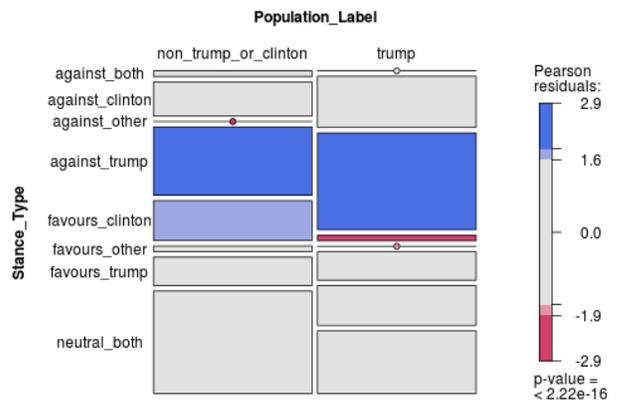

**Figure 3:** Trump Stance Type Spotlight – Residuals Mosaic

Moreover, beyond *stance types*, we also see that, compared to any other population, the *attr_type* of Clinton attributions was more often a *political_platform*, whereas Trump attributions more often involved calling out a specific group, à la *group_callout* (e.g. Mexico/China). While the sheer growth of feature combinations complicates grphical analysis of *stance_type * attr_type* and *stance_type * cue_valence* interactions, we do intuitively tend to see *positive* cue valences associated to favourable stances, *negative* cue valences associated to unfavourable stances, and *neutral* stances most often associated to *callout* attribution types (as opposed to *personal stances*).

## 4.2 Identifying Inherent Challenges at Scale

Unfortunately, several uncovered systemic challenges limit our immediate ability to scale and extend this approach. Most obviously, enforcing a threshold of inter-annotator labelling agreement and crowdsourcing beyond a small cohort of annotators would achieve replicability over far more attributions than the collective consensus currently required for our pilot round.

Yet, even in our pilot data, inherent obstacles inhibit our ability to draw conclusions. For the purposes of future research, we now explicitly enumerate these challenges and demonstrate positive results towards their resolution.

(1) Avoiding Bias in our Attribution Samples

(2) Identifying the Correct Attribution Source

(3) Limiting Cascading Errors

To understand the first problem, it is best to illustrate the breakdown of our labelled attributions:

| Publication_Label | Num_Articles | Count_Trump | Count_Clinton | Count_Other |
|---|---|---|---|---|
| breitbart | 3 | 11 | 28 | 45 |
| huffpost | 3 | 11 | 10 | 29 |
| nyt | 5 | 23 | 11 | 70 |
| politico | 4 | 40 | 33 | 64 |
| usa-today | 4 | 11 | 10 | 27 |
| wash-post | 3 | 19 | 11 | 94 |
| west-journal | 4 | 8 | 4 | 36 |
| | | | | |
| Totals | 26 | 121 | 100 | 365 |

**Table 2:** Breakdown of our Labelled Attributons

Immediately, it is easy to see that, despite collecting at least 100 attributions each for Hilary Clinton and Donald Trump and totaling over 600 unique attributions across over 25 articles, this effort ultimately breaks down to only around 10 Trump and 10 Clinton attributions per media outlet. Moreover, even after annotating around 50 attributions across 4 distinct articles, we were unable to gather even 10 Clinton or Trump attributions from West-Journal.

Seeking to scale beyond 3-5 articles per publisher, it then becomes unclear whether it will be possible to satisfy both (i) a relatively equal distribution of articles per publisher, and (ii) a sufficiently large number of Trump and Clinton attributions per media outlet. Fortunately, our pilot data has been effective in taking first steps to resolve this problem.

While problem (1) may hinder the extensibility of our partisan trait comparisons, it does not reduce confidence that our sample data can act as a reasonably large validation set of Trump and Clinton attributions distributed across several media outlets. Thus, the sample data's immediate usefulness arises in developing and validating an attribution-source classification algorithm for future reuse in identifying [and understanding the scale of] Trump and Clinton attributions across our publishers.

Evidently, problem (2) then arises in developing this classifier. Soon, we will indeed demonstrate some success in implementing said classification algorithm. Before doing so, however, we address the challenges observed by our baseline model leveraging state-of-the-art (Lee et al., 2013) co-reference resolution:

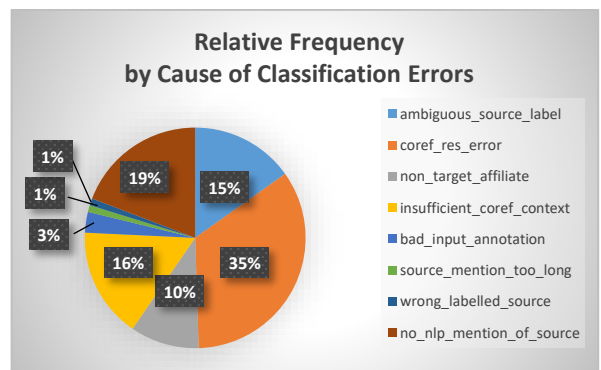

**Figure 4:** Errors in Identifying the Attribution Source

Elaborating on the above source identification errors, we see that the most frequent bin (35%) was due exclusively to inadequate performance of vanilla co-reference resolution. This is not too surprising given that the baseline algorithm had indeed advertised F1 scores typically in the range 0.65-0.75 out of 1. More troubling, however, are the remaining cases. To elucidate ambiguous source labels (15%), consider the source "Trump Campaign Spokesman, Hope Hicks". Do we label this as "Trump"? What about just "The Trump Campaign" or "An insider close to Trump"?

Beyond ambiguouity, other inherent issues include tricky non-target sources (e.g. "A 2008 Clinton Veteran"), correct but inadequate co-reference resolution (e.g. "he" -> "the leader"), and other logistical errors (introduced by various human errors, and programmatic limitations).

Finally, human judgement lies at the root of both problem (3). Regarding cascading errors, consider the following attribution: "**Clinton** refused to offer a mea culpa for *setting up a private server and sending around sensitive government emails*". Arguably, this is not an attribution at all since it ascribes only a nullity of action to Clinton, failing even to assign her a positive, negative, or neutral stance towards the email dilemma. On the other hand, it is somewhat clear, at a microscopic level, that the statement, as evidenced primarily by "refused to offer a mea culpa for" opposes Clinton's handling of the email situation. Thus, in this case, the annotator & reviewer agreed in assigning this as an attribution sourced to Clinton with a stance type *against_clinton*. In this case, it probably would have been best to exclude attributions of this nature in the BRATT file for this article. From this example, it is easy to see how the early & intermediate reliance of our research on human annotation can cascade into subtle errors in later phases.

### 4.3 Attribution Source Classification

Fundamental in our efforts to resolve problems (1) and (2) as described in the last section, solid progress has been made towards computationally classifying attributions as one of: {Donald Trump, Hilary Clinton, Other}. Specifically, the model we present was found to perform with accuracy greater than 88%, and (precision, recall) ≈ (95%, 80%), in binarized labelling targeted at either candidate.

In brief, the algorithm that was found to perform best primarily leverages state-of-the art co-reference resolution (Lee et al., 2013) as a baseline, with a few simple but key optimizations. Summarizing, our baseline model relies on vanilla co-reference resolution to extract the *most-representative* mention of a given source text (e.g. "he" -> "Donald Trump"), before proceeding to match on the result via custom regular expressions. The main problem with this baseline model is that it can often (≈ 19% of the time) fail to extract *any* representative mention at all. Fortunately, the cause of this is eminently fixable in that it manifests most often when the source text was annotated too precisely (e.g. "hopeful future president Mr. Donald John Trump"). In which case, if we fail to extract a *most-representative* mention, we see big improvements by defaulting to match against the source text itself. Note, however, that we do prefer to match the *most-representative* mention when possible in case, for example, "Clinton" maps to "Bill Clinton" (in which case, we do not label Hilary Clinton). In a similar frame of mind, empirical observation found that indeed long source mentions almost always mean trouble for our matchers (e.g. "Hilary Clinton, wife of former president Bill Clinton"). Here, an obvious improvement would be to extract and operate only on the sentence *subject* but, for now, we found that it was sufficiently effective to only ever operate on the first 5 tokens of an attribution's source text field. Lastly, in optimizing our performance, it was essential to ensure our regular expressions were both exclusive enough to handle frequent error cases (e.g. "Bill Clinton", "The Clinton Administration", "Donald Trump Jr.") but permissive enough to capture most valid instances of "Trump", "The Clinton Campaign", and other expected variations.

## 5 Discussion

Reviewing the results, despite some emergent challenges and surprises in our preliminary statistical analysis, we arrive at many actionable insights. Our pilot data has displayed both favorable and discouraging characteristics but illuminated many directions to progress. In addressing some of the emergent challenges, our source classification algorithm has shown promise as an immediate first step in future research.

On retrospective, we are initially discouraged in that many of the features we originally envisioned as important failed to convey evidence of significant contribution in differentiating our populations of interest. Namely, none of *source_valence, is_direct_quote*, or *is_tweet* ever showed evidence of contribution, and *honorific_text* only seemed to assist in differentiating Trump from Clinton based on a select few instances of "Mr." vs. "Mrs.". Yet, *stance_type* frequently demonstrated evidence for significance, and often in tandem with *cue_valence*, and *attr_type*. In this respect, we are encouraged to further explore how these features interact to convey partisan characteristics. For instance, we might wonder if perhaps *cue_valence* acts a modifier for the attributed stance, such that a stance type of *favours* or *against* could both potentially portray the stance's **source** either *positively* or *negatively* depending on the associated cue. Similarly, based solely on these features, we might investigate trends in the way sources are attributed to speaking of themselves, compared to how they speak of others, and how others speak of them.

Furthermore, we should pay heed to the error sources we have identified in scaling our approach. Our ability to draw partisan inter-publisher comparisons is currently hindered by inadequacies in the magnitude of our per-publisher datasets. In fact, it is not yet entirely clear whether it will even be possible to extensively expand our data in an unbiased manner. Fortunately, in addressing this concern, we have had success thus far in developing a well-performing classifier to facilitate scaling our dataset pivotal on attribution sources. We have also recognized that it will be especially important to consider cases of *ambiguity* and *human subjectivity* moving forward.

Of most immediate future research directions, there is considerable opportunity to incorporate supervised learning approaches in this domain. Directly targeting political bias, we might simply approach this as seeking to predict the *news outlet* based on our perceived partisan features (e.g. *cue_valence*, *stance type*, etc.). Alternatively, we might instead design some *political bias* metric and seek to measure how it varies as a function of our news publisher. From another perspective, it is also worth investigating the partisan contribution of non-categorical attribution features (e.g. a lemmatized bag of words for the cue, source text encoding, etc.). It may even be possible to accurately predict some of our classification labels, without the need for human annotation, indirectly from these implicit features.

## 6 Conclusion

In this paper, we have given formal treatment to the intuitive notion of partisanship in journalistic attributions. Here, we operationalize and develop an approach to statistically characterizing the partisan traits of these attributions and demonstrate success in validating this approach. Moreover, we identify challenges to address in further extending our work and begin to implement the framework to progress in this domain. We make available our labelled pilot attribution data, and attribution source classifier for this purpose.

Moving forward, this classifier shall be immediately applicable in amassing further attributions sourced to Hilary Clinton and Donald Trump drawn from articles published by several distinct media outlets. On this dataset, it will then be possible to better analyze partisan traits in our data at scale and explore the effectiveness of machine-learning approaches in characterizing these implicit attribution features.